\documentclass[journal]{IEEEtran}
\usepackage{url}
\usepackage[pdftex]{graphicx}
\usepackage{csquotes}
\usepackage{acronym}
\usepackage{multirow}
\usepackage{adjustbox}
\usepackage{amsmath}
\usepackage{subcaption}
\usepackage{filecontents}
\usepackage{colortbl}
\usepackage{textcomp} % for arrow in text
\usepackage{enumitem}
\usepackage[ruled]{algorithm2e}
\usepackage{threeparttable}
\usepackage{tablefootnote}
\usepackage{lipsum}
\usepackage{todonotes}
\usepackage[mathlines,switch]{lineno}
\usepackage{pifont} % http://ctan.org/pkg/pifont
\usepackage[english]{babel}
\usepackage{tikz}
\usepackage{setspace}
\usepackage{soul}
\usetikzlibrary{positioning}
\usepackage{cite}
\usepackage{etoolbox}
\makeatletter
\patchcmd{\@bibitem}{In}{\relax}{}{}
\makeatother

\begin{document}

\title{Quality Prediction of AI Generated Images and Videos: Emerging Trends and Opportunities}
\author{
    \IEEEauthorblockN{Abhijay Ghildyal~\IEEEmembership{Student Member,~IEEE}\IEEEauthorrefmark{1}, Yuanhan Chen\IEEEauthorrefmark{2}, Saman Zadtootaghaj\IEEEauthorrefmark{3}, Nabajeet Barman~\IEEEmembership{Senior Member,~IEEE}\IEEEauthorrefmark{4}, Alan C. Bovik~\IEEEmembership{Life Fellow,~IEEE}\IEEEauthorrefmark{5}} \\
    \IEEEauthorblockA{\IEEEauthorrefmark{1}Portland State University, abhijay@pdx.edu} \\
    \IEEEauthorblockA{\IEEEauthorrefmark{2}Sony Interactive Entertainment, Aliso Viejo, USA, May.Chen1@sony.com} \\
    \IEEEauthorblockA{\IEEEauthorrefmark{3}Sony Interactive Entertainment, Berlin, Germany, Saman.Zadtootaghaj@sony.com} \\
    \IEEEauthorblockA{\IEEEauthorrefmark{4}Sony Interactive Entertainment, London, United Kingdom, Nabajeet.Barman@sony.com} \\  
    \IEEEauthorblockA{\IEEEauthorrefmark{5}Department of ECE, University of Texas at Austin, Austin, TX, USA, bovik@ece.utexas.edu}   

\thanks{Abhijay Ghildyal is with Portland State University, Portland, USA. Yuanhan Chen, Saman Zadtootaghaj and Nabajeet Barman are with Sony Interactive Entertainment (Aliso Viejo, US; Berlin, Germany and London, UK respectively). Prof Alan C. Bovik is with Department of Electrical and Computer Engineering, The University of Texas at Austin, Austin, TX, USA.
\newline
Corresponding Author: Nabajeet Barman (n.barman@ieee.org)}}

\markboth{Preprint of a manuscript currently under review}
{Shell \MakeLowercase{\textit{et al.}}: Bare Demo of IEEEtran.cls for IEEE Journals}

\maketitle

\begin{abstract}
The advent of AI has influenced many aspects of human life, from self-driving cars and intelligent chatbots to text-based image and video generation models capable of creating realistic images and videos based on user prompts (text-to-image, image-to-image, and image-to-video). As these new AI models mature, their accuracy and functionality will make them integral to many real-world services and products. One field that is being revolutionized by AI-based techniques is multimedia, impacting everything from content generation and encoding to packaging, delivery, and display on end-user devices. AI-based methods for image and video super resolution, video frame interpolation, denoising, and compression have already gathered significant attention and interest in the industry and some solutions are already being implemented in real-world products and services. However, to achieve widespread integration and acceptance, AI-generated and enhanced content must be visually accurate, adhere to intended use, and maintain high visual quality to avoid degrading the end user’s quality of experience (QoE). 

One way to monitor and control the visual ``quality'' of AI-generated and -enhanced content is by deploying Image Quality Assessment (IQA) and Video Quality Assessment (VQA) models. However, most existing IQA and VQA models measure visual fidelity in terms of ``reconstruction'' quality against a pristine reference content and were not designed to assess the quality of ``generative'' artifacts. To address this, newer metrics and models have recently been proposed, but their performance evaluation and overall efficacy have been limited by datasets that were too small or otherwise lack representative content and/or distortion capacity; and by performance measures that can accurately report the success of an IQA/VQA model for ``GenAI''. This paper examines the current shortcomings and possibilities presented by AI-generated and enhanced image and video content, with a particular focus on end-user perceived quality. Finally, we discuss open questions and make recommendations for future work on the ``GenAI'' quality assessment problems, towards further progressing on this interesting and relevant field of research.
\end{abstract}


\section{Introduction}

The advancing fields of machine learning and computer vision are constantly changing as new tasks and challenges continuously emerge, driven by advancements in technology and expanding application areas. As researchers develop innovative methods to tackle these new challenges, being able to demonstrate their effectiveness becomes increasingly crucial. However, establishing standardized approaches for evaluation remains a complex endeavor. One significant domain where computer vision is making a significant impact is the field of multimedia communications. Machine learning is revolutionizing various processes in this field, including signal acquisition, to compression, streaming, analysis, and enhancement, resulting in more intelligent, efficient and enhanced user experiences. One very recent task having great promise is content generation using advanced deep models such as GANs, VAEs, LLMs and more recently, Diffusion Models. AI generated text, audio, images and video content are typically referred to as AI Generated Content (AIGC).

A critical aspect of any video communication system is the ability to be able to measure and evaluate end user Quality of Experience (QoE). Unlike traditional technical quality of service factors, QoE encompasses the subjective perceptions and experiences of the users as they interact with different multimedia content such as audio, images and videos. The success of any service or application is utlimately dependent on its acceptance by the end user. Hence, understanding and measuring end user QoE is imperative, as it reflects the degree of satisfaction of the end-users. Objective models and metrics that can measure end user QoE are indispensible for the optimization and advancement of numerous applications. Innovations that can similarly be used to measure and optimize the delivery of AI-generated content are highly desirable, given future expectations that ``Gen AI'' will pervade streaming media, television, gaming, advertisements, and more. 

\begin{figure*}[t]
\scriptsize
\centering
\includegraphics[width=\textwidth]{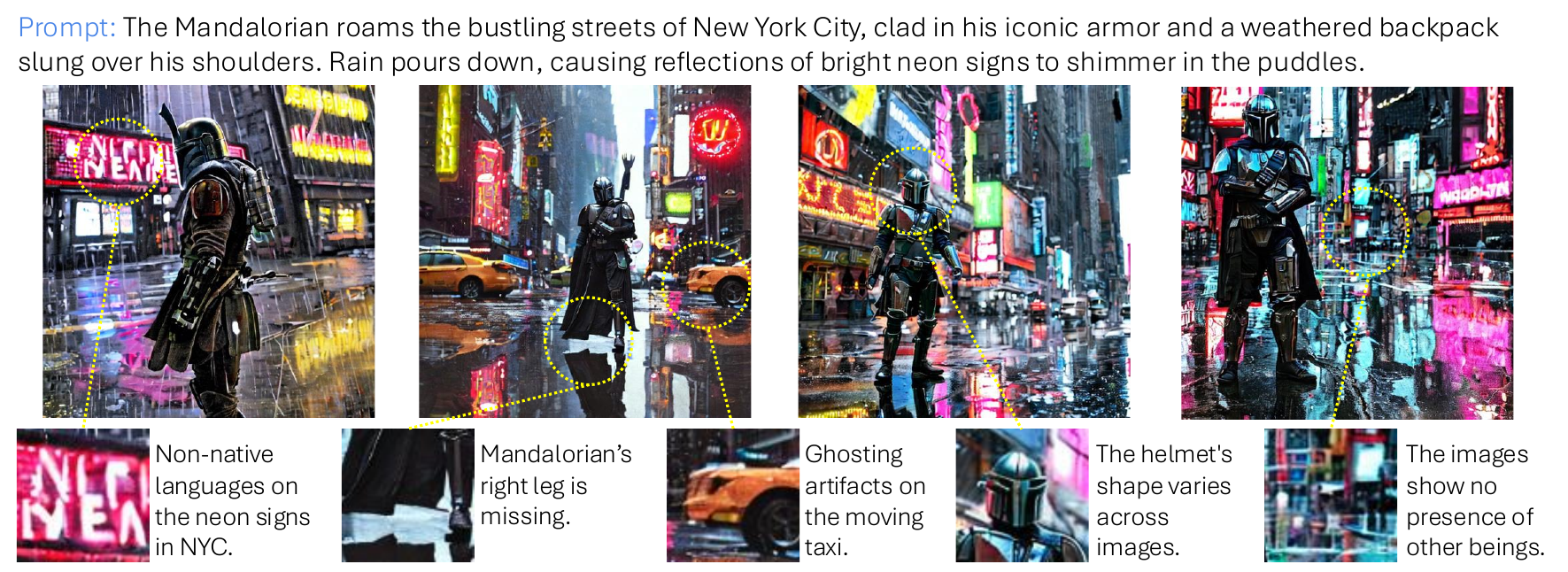}
\caption{Four output images generated by the Text2Image Stable Diffusion model based on the provided prompt. The lower section highlights missing or incorrect components in the generated images, underscoring the need for a quality assessment model that can holistically evaluate both technical aspects and the model's understanding of common sense.}
\label{fig:common_sense}
\end{figure*}

As Generative AI models continue to transition from research to industrial and consumer applications, ensuring the quality of generated images and videos is becoming crucial. Humans have an exceptional ability to assess visual quality, and assuming normal vision. They are sensitive not only to details such as image sharpness, color accuracy, and the presence of distortions, but also to the realistic portrayal (or lack there of) of objects and scenes. For instance, when viewing generated content, a human can easily discern whether it appears natural or if there are anomalies such as unnatural lighting or distorted textures. 

The most accurate way to assess end-user QoE is to ask a set of human subjects to supply opinions of whichever aspects of quality are of interest. However, recruiting human subjects is not cost-effective and in many cases, is not practical, e.g. for the QoE prediction of massive quantities of content, or for real-time quality assessment. Automated quality prediction models such as SSIM~\cite{SSIM} and VMAF~\cite{NetflixVMAF_Github} exemplify how such models can be successfully deployed at the largest scales. 

However, AI generated content brings new artifacts that express different aspects of perceptual, technical, and aesthetic quality. For example, Fig.~\ref{fig:common_sense} shows output images from a Text-2-Image AI model which may be aesthetically pleasing, but lack adherence to the input text prompt and often fail to express real-world knowledge. Existing IQA/VQA methods are not capable of capturing such artifacts (see Section~\ref{subsec:challIQAVQA}), and hence generally do not suffice for predicting the quality of content produced in Generative AI workflows. This limits their usefulness for GenAI model design, workflows, and system evaluation. Generally, there is a significant need for further research on the development of both IQA and VQA models tailored specifically to address the challenges posed by Generative AI technologies. 

Towards advancing progress on these problems, we conduct an in-depth discussion on the necessity of evolving image and video quality metrics to meet the demands of the increasingly complex and diverse tasks involving AI generated and modified content. We discuss the recent advancements in these directions, and identify unsolved challenges. We explain in greater depth the need to innovate GenAI specific image and video quality models and why this is difficult; and why it is important to curate new datasets of GenAI content labelled with human with subjective ratings.

The rest of the paper is organized as follows. Section~\ref{sec:background} provides an overview of the various types of GenAI content along with emerging challenges in the field of GenAI Image Quality Assessment (IQA) and Video Quality Assessment (VQA). Section~\ref{sec:aigc_aiec} delves into a detailed discussion of AI-Generated Content (AIGC) and AI-Enhanced Content (AIEC). Section~\ref{sec:QA} surveys both objective and subjective quality assessment methods, and recent work on the quality assessment of AIGC and AIEC. Section~\ref{sec:challenges} addresses key challenges and opportunities that the research community must tackle to further advance the field. Section~\ref{sec:trends} highlights recent trends in IQA and VQA for AI-generated and AI-enhanced content, providing readers with insights into potential directions for future research. Finally, Section~\ref{sec:conc} concludes the paper.

\section{Background} \label{sec:background}

\subsection{Content Types}
Over the past few decades, multimedia content has significantly evolved, transitioning from the low-resolution analog television of te 1980s to today's ultra high-definition videos captured using professional-grade devices and displayed on 4k (or larger) displays. Advances in technology have enabled the production of high-quality content that is largely free of visual artifacts. This category of signal is commonly referred to as professionally generated content (PGC). Examples include movies and television shows produced by large production studios such as Universal Studios\footnote{\url{https://www.universalstudios.com}} and Sony Pictures Entertainment\footnote{\url{https://www.sonypictures.co.uk/}}, and photographs captured by professional photographers. Over the past two decades, other forms of non-2D content, such as free-viewpoint televisions~\cite{tanimoto2010free}, light field imaging~\cite{8022901, 9175517}, point clouds~\cite{liu2023point, yang2022no}, and 3D meshes~\cite{7532509} have also become quite popular. These technologies go beyond traditional images and videos to capture and represent depths, shapes, and larger dimensions to provide end users more immersive experiences. 

Recently, social media platforms have become integral to modern communication and entertainment, revolutionizing the way people create, share, and consume content using consumer-grade devices. Given the pervasive availability of mobile devices such as smartphones and body-worn cameras like Go-Pro, \textit{User-Generated Content (UGC)} has become increasingly popular as evidenced by the deployment of massive platforms such as YouTube, Instagram, and TikTok. 

More recently, there has been a rise in \textit{AI-Generated Content (AIGC)} platforms such as Sora,\footnote{\url{https://openai.com/index/sora/}} Synthesia,\footnote{\url{http://synthesia.io}} and Midjourney.\footnote{\url{https://www.midjourney.com/home}} These platforms leverage newer AI technologies such as GANs and Diffusion models to create increasingly realistic or artistic multimedia content. Users can input their requirements media desires as text, or as an initial image to guide the AI in generating customized content tailored to their specifications. These advancements represent a significant evolution in digital media, showcasing the practical capabilities of GenAI in creative contexts. There is also \textit{AI-Enhanced Content (AIEC)}, which refers to media that has been modified or improved through the use of AI methods. This includes enhanced content creation, such as improving image quality, synthesizing photorealistic content, automating aspects of video editing, and modifying parts of content based on user prompts, among a great many other possibilities.

Since GenAI is an evolving field, different terminologies are used across the literature. In the absence of standardized definitions and for an easier understanding of this paper and the topic under discussion, we outline commonly used terminologies throughout the paper:

\begin{itemize}
    \item \textit{AIGC} or AI generated content refers to content \textbf{generated in its entirety} by an AI algorithm based on an input prompt, which can be text, image, video, sound, or a combination of these elements (see section III-A).
    \item \textit{AIEC} refers to any \textbf{existing content} that has been enhanced or modified by an AI algorithm, which might in the process include generating a part of the content.
    \item \textit{Natural content} refers to pictures or videos of the real world that are captured with photographic or videography cameras.
    \item Graphical content refers to computer animations not created using GenAI.
    \item Traditional IQA/VQA methods refer to models and algorithms such as PSNR, SSIM, VMAF, and UVQ~\cite{uvq} which are designed for traditional UGC and/or PGC content.
\end{itemize}
Of course, some content may contain more than one of these, such as screen content. Details and discussions of these terms may be found in Section~\ref{sec:aigc_aiec}.

\subsection{Recent Challenges in IQA and VQA} \label{subsec:challIQAVQA}

Despite rapid advancements of AI technologies that are finding their way into media products and services, the ultimate success of GenAI content will depend on its acceptance by end users. Content modified or generated by AI algorithms must meet users' requirements and expectations, and provisions and technologies must be made to ensure that they are fair, unbiased, and respect the privacy and copyrights of other users and content creators.

One of the ways to measure end-user QoE is via image and video quality assessment models and algorithms that typically measures the ``technical'' or ``reconstruction quality'' of content. Over the past two decades, numerous image and video quality assessment methods have been developed and deployed in practical media workflows. However, IQA and VQA models designed for natural content are not suitable to assess the perceived (subjective) quality of AIGC. There are several reasons for this, among which is the fact that GenAI content may not obey statistical image models that reliably characterize natural content~\cite{lee2024holistic}, and since AI-generated Content may produce artifacts, such as extra limbs or impossible animals, that do not fall into the usual understanding of ``distortions.''

\begin{table*}[t!]
\centering
\caption{Example Datasets used for Training and Evaluation of AI Generated and Enhanced Content.}
\label{table:summary}
\resizebox{1.0\textwidth}{!}{%
\begin{tabular}{|>{\raggedright\arraybackslash}p{2.5cm}|>{\raggedright\arraybackslash}p{3cm}|>{\raggedright\arraybackslash}p{2.5cm}|>{\raggedright\arraybackslash}p{5cm}|>{\raggedright\arraybackslash}p{3cm}|}
\hline
\textbf{Database} & \textbf{Type} & \textbf{Size} & \textbf{Application} & \textbf{Models} \\ \hline
WebText ~\cite{radford2019language} & Text & 40 GB, 570 GB & Large Language Model Training & GPT-2~\cite{radford2019language}, GPT-3~\cite{brown2020language}, CTRL~\cite{keskar2019ctrl} \\ \hline
BookCorpus~\cite{zhu2015aligning} & Text & 6 GB, 11,038 books & Long-range text dependencies & BERT~\cite{devlin2019bert}, RoBERTa~\cite{liu2019roberta}, XLNet~\cite{yang2019xlnet} \\ \hline
COCO (Common Objects in Context)~\cite{lin2014microsoft} & Image & 330K (\textgreater 200K labeled) & Image Captioning, Object detection & Pix2Pix~\cite{isola2017image}, VQ-VAE-2~\cite{razavi2019generating}, CLIP~\cite{radford2021learning} \\ \hline
ImageNet~\cite{deng2009imagenet} & Image & 14 million & Benchmarking Image Classification & BigGAN~\cite{brock2018large}, StyleGAN~\cite{karras2019style}, DALL-E~\cite{ramesh2021zeroshot} \\ \hline
Conceptual Captions~\cite{sharma2018conceptual} & Text - Image & 3.3 million & Image Captioning, Multimodal Learning, Content Generation & CLIP~\cite{radford2021learning}, VisualBERT~\cite{li2019visualbert}, Oscar ~\cite{li2020oscar}\\ \hline
CLIP (LAION-400M)~\cite{radford2021learning} & Text - Image & 400 million image-text pairs & Creative AI, Content Moderation & CLIP~\cite{radford2021learning}, DALL-E~\cite{ramesh2021zeroshot} \\ \hline
LLFF (Local Light Field Fusion) Dataset & Images - Corresponding camera parameters & 5-20 GB & Novel View Synthesis, Neural Rendering, 3D Reconstruction & NeRF~\cite{mildenhall2020nerf}, NSVF~\cite{liu2020neural}, DeepVoxels~\cite{sitzmann2019deepvoxels}\\ \hline
Tanks and Temples Dataset~\cite{knapitsch2017tanks} & 3D models & 10 - 100 GB & 3D reconstruction validation & NeRF~\cite{mildenhall2020nerf}, NSVF~\cite{liu2020neural}, DeepVoxels~\cite{sitzmann2019deepvoxels} \\ \hline
DIV2K~\cite{agustsson2017ntire} & Images & 7 - 8 GB & Super Resolution & ESRGAN~\cite{wang2018esrgan}, SRGAN~\cite{ledig2017photo}, EDSR~\cite{lim2017enhanced} \\ \hline
Vimeo-90K~\cite{xu2018vimeo} & Videos & 82 GB & Super Resolution, Frame Interpolation, Video Denoising & Super Slomo~\cite{jiang2018super}, BasicVSR~\cite{chan2021basicvsr} \\ \hline
UCF-101~\cite{soomro2012ucf101} & Video & 13,320 Videos, 27 Hours & Action recognition, video Prediction, Frame Interpolation & I3D~\cite{carreira2017quo}, MoCoGAN~\cite{tulyakov2018mocogan}, DPC~\cite{liu2019dpc} \\ \hline
\end{tabular}}
\vspace{-0.05in}
\end{table*}

The growing use of AI to enhance, modify, or generate content (image, video, audio, speech) will necessitate the formulation of new conceptions of ``visual quality'' as GenAI models introduce ``generative" distortions, the appearances of which do not comport with commonly encountered video degradations. For example, Fig.~\ref{fig:common_sense}, depicts several images generated by Stable Diffusion~\cite{rombach2022high} using the prompt given in the Figure. It may be observed that the generated images appear to be of reasonably high quality in the usual sense, including impressive reflections in the puddles on the street. However, there remains substantial room for improvement in the model's ability to integrate world knowledge, as is evident upon a close examination of the image.

For instance, despite the prompt mentioning a \textit{weathered backpack}, it is visible in all the images. Although the prompt specifies that the character should be roaming bustling streets, there are no other beings present. The setting appears to be New York City, resembling Times Square, yet the language on the neon signs is nonsensical, unlike any language in the Ethnologue. Moreover, the character is positioned in the middle of the street, whereas it would be more realistic for them to be walking on the sidewalk amidst a crowd of people. Evaluating AI-generated content is challenging, as it involves not only assessing technical quality, but also judging common sense, semantic accuracy, and composition. Furthermore, evaluating a model's ability to spatially integrate real-world knowledge is significantly more complex than assessing the technical quality of its outputs. Therefore, this indicates the need for GenAI content evaluation models that are able to measure not only technical quality, but also other aspects such as alignment with the input prompts, semantic correctness, biases, and aesthetics, among others.

\section{AI Generated and Enhanced Content} ~\label{sec:aigc_aiec}

The field of AIGC encompasses a wide range of modalities including images, videos, 3D meshes, point clouds, NERFs, and many others. Here we limit our discussion t images and videos, Although many of the arguments and insights may hold for other modalities as well.

Next we discuss AI Generated content (AIGC) followed by AI Enhanced Content (AIEC). Table~\ref{table:summary} presents a summary of key databases uses for AIGC and AIEC training and testing, including their types, sizes, and associated models. Each dataset is accompanied by details on content, target application, and examples of models trained or evaluated on it. These datasets are critical in advancing research in areas such as large language model training, image captioning, 3D reconstruction, and more. 

\subsection{AIGC}
Generative Artificial Intelligence (GenAI) is a type of AI that accepts text, images, audio, and videos as input, then uses it to create new content. AIGC refers to content fully generated by AI models, in response to human prompts or requests. AIGCs are being used in a wide variety of applications such as Content Creation, Marketing and Advertising, Entertainment, Education, and Healthcare, among many others~\cite{zhou2023aigc}. Next we briefly discuss contemporary image and video generation modalities.

\subsubsection{Image Generation}
Since the early famous introduction of StyleGANv2 with its ``This person does not exist"~\cite{karras2020analyzing}, visual content generation has significantly evolved in recent years. GANs, variational autoencoders (VAE), and diffusion models have all significantly influenced the evolution of image and video generation models~\cite{ho2020denoising}. Generative image models can be conveniently divided into several categories:
\begin{itemize}
    \item \textit{Text2Image}:
A text-to-image model is an AI-based generative model that synthesizes images in response to textual prompts. These systems combine language models for input text transformation and generative image models with content alignment. Notable instances of text-to-image models include DALL-E, Midjourney, etc.~\cite{ramesh2021zeroshot}. 
    \item \textit{Image2Image}:
Image-to-image generation models modify images with target tasks such as image enhancement, denoising, deblurring, or style transfer. Some of the most significant and popular image-to-image models are Stable Diffusion and Pix2Pix~\cite{rombach2021highresolution,isola2017image}.
\item \textit{Hybrid Approaches} More recently, hybrid models have been developed that combine various generative architectures to produce content based on both visual and textual information, including DALL-E, VQ-VAE-2, and StyleGAN with Condition Inputs. 
\end{itemize}

\subsubsection{Video Generation}
More recently, there has been remarkable progress on the more complex, higher-dimensional and compute-intensive problem of generating visual content using machine learning. Generative video generation involves the use of artificial intelligence techniques to create realistic (and/or artisitic) videos based on learned patterns from existing video data. This field extends generative AI image generation to the temporal domain, leveraging deep learning models to generate dynamic sequences of frames that resemble real-world (or possibly other-worldly) videos.

\begin{itemize}
    \item Text-to-Video: Text-to-video (T2V) involves generating sequences of images based on input text descriptions. Applications such as Sora by OpenAI typically use transformer architectures that operate on spacetime patches of video and image latent codes, and are capable of generating short, high fidelity videos~\cite{chen2024sora}. “Make-A-video” by Meta AI extends diffusion-based T2I advancements to T2V~\cite{singer2023makeavideo}. Text inputs pass through a decoder to create image embeddings, which are then interpolated over time at a special framerate, yielding resulting output video.
    \item Image-to-Video: Image-to-Video generative models typically accept an image or a series of images as input,  generating output video sequences by predicting future frames or by generating frames between existing frames in a coherent and consistent manner. Notable example including MocoGAN~\cite{tulyakov2018mocogan}, Vid2Vid~\cite{wang2018vid2vid}, and TGAN~\cite{saito2017temporal}.
    \item Video-to-Video: Recent progress has been made on mapping videos to videos to achieve resolution enhancement and frame interpolation~\cite{Liang_2024_CVPR,rerenderavideo,motioni2v,fairy}.
\end{itemize}

\subsubsection{Novel View Synthesis}

Novel view synthesis methods such as Gaussian Splatting and Neural Radiance Fields (NeRF) aim to generate new, realistic and consistent viewpoints of a scene using information from a given set of images. Gaussian Splatting leverages a probabilistic model to render novel views by ``splatting" multiple Gaussian functions of different spreads and amplitudes over a scene. NERFs, rely on neural networks to learn volumetric scene functions, while capturing relevant lightning effects and associated scene details. Quality assessment methods and models able to capture the ``quality" in terms of the strengths and weaknesses of these new methods could help advance these technologies further. However, as mentioned earlier, we limit our discussion to the processing of 2D images and videos.

\subsection{AIEC}

In addition to the emerging field of AI generative content, a more mature field is AIEC where existing image and video contents are ``enhanced" or ``modified" using AI-based methods such as super resolution~\cite{super_res_survey}, denoising, deblurring, compression artifact removal, inpainting, (video) frame interpolation (VFI), among many others~\cite{dl_image_processing}. AIEC methods focuses on enhancement of the (visual) quality and appeal of \textbf{existing} image and video content. 

Image and video enhancement methods involve using algorithms to restore or enhance visual quality, making the image clearer, of better contrast, or more detailed, or closer in appearance to a reference image~\cite{YE2024102614}\cite{GUO2023e14558}\cite{LIU2022103547}. Poor visual quality can arise from various factors, such as unnatural variations in brightness or color, blurriness caused by camera shake, out-of-focus capture, subject motion, lower resolution or insufficient pixel density, and excessive compression.

Methods deploying, deep learning networks such as CNNs, GANs, and Autoencoders have significantly advanced these and other types image reconstruction. Networks for image reconstruction typically learn to optimize recreating pixels against reference images during training. However, succeeding of these tasks remain challenging, especially for unforeseen and out-of-distribution data. Therefore, there is a growing interest in recent developments in generative models, owing to their capability to generate high-quality content~\cite{saharia2022image}\cite{li2023diffusion}. AIEC then involves learning a generative component that enhances images while maintaining fidelity to ground-truth. While excellent results have been obtained on a wide variety of AIEC tasks, little attention has been applied to objectively evaluating the perceptual outcomes of these models.

It is worth mentioning in passing that while there is generally a clear distinction between AIGC and AIEC technologies, in some cases, such as VFI or NERFs, they might be considered as either AIGC or AIEC. We will discuss any ambiguous interpretations as they arise.

\section{Quality Assessment} \label{sec:QA}

\subsection{Definition of Quality} 

``Quality,'' in the context of Image Quality Assessment (IQA) and Video Quality Assessment (VQA), refers to the perceived visual clarity and degree of visual distortion (or lack thereof) of images and videos as judged by human observers (``subjective quality'') or predicted using perceptual image models and /or machine learning (``objective quality''). Quality assessment can be affected by various attributes such as contrast, sharpness, color accuracy, and the presence of acquisition or processing distortions. Acquisition distortions may include various types of blur, sensor noise, over/under exposure, and so on. Processing distortions can include blocking from compression, blur or ringing from scaling, frame-rate alterations, among many others.

\subsection{Objective Image and Video Quality Assessment} \label{sec:objective_iqa_vqa}
Objective Image and Video Quality Assessment methodologies refer to models and algorithms that assess image/video quality based on quantitative perceptual models make it possible to obtain consistent and repeatable predictions of various aspects of media quality. Accurate visual quality predictors are widely used to assess the perceptual quality of images and videos in large-scale industry streaming and social media workflows. Such objective IQA and VQA approaches can broadly be classified into three categories: Full-Reference (FR), Reduced-Reference (RR), and No-Reference (NR). FR methods require access to the full source/reference sequence as they measure the visual signal or information fidelity of the test sequence as compared to the reference sequence. RR methods estimate the amount of distortion in the distorted sequence using only partial information from the original sequence. The subset information could vary from motion vectors and quantization parameter (QP) values, to texture. By contrast, NR models estimate various distortions present in the image or video sequence and quantify their impact on subjective quality. NR models have gained much attention due to their suitability for assessing quality of in-the-wild media uploaded to social media applications, and in live streaming and cloud gaming applications where there is no reference signal available.

\subsubsection{Traditional Methods and Models}

The most widely used IQA and VQA model employ features derived from mathematical models of distortion perception. Several models that are widely used and marketed by industry rely on neurostatistical models of distortion perception, viz. they model perturbations of the responses of visual neurons to the presence of distortion. These include VIF~\cite{vif}, NIQE~\cite{mittal2012making}, BRISQUE~\cite{mittal2012no}, and VMAF~\cite{NetflixVMAF_Github}. Other widely used models include the antiquated (non-perception based) PSNR (Peak Signal-to-Noise Ratio)\cite{wang2009mean} and the Primetime Emmy-winning SSIM (Structural Similarity Index)~\cite{SSIM} measure pixel-level differences squared differences and structural distortions between a reference and a test image or video, respectively. VMAF~\cite{NetflixVMAF_Github} fuses several VIF~\cite{vif} features with change and detail features using an SVR. While these modes are effective in many practical scenarios, they are limited in their ability to handle complex combinations of real-world distortions, and they are unable to account for high-level semantic visual aspects that also contribute to perceived visual quality.

In recent years, a variety of deep learning-based approaches such as LPIPS~\cite{kettunen2019lpips}, PieAPP~\cite{Prashnani_2018_CVPR}, DISTS~\cite{ding2020image}, ST-LPIPS~\cite{ghildyal2022stlpips}, ARNIQA~\cite{agnolucci2024arniqa}, UNIQA~\cite{yun2023uniqa}, NDNetGaming~\cite{NDNetgaming}, UVQ~\cite{uvq}, PaQ-2-PiQ~\cite{ying2020patches}, RAPIQUE~\cite{Rapique}, CONTRIQUE~\cite{contrique}, and Re-IQA~\cite{ReIQA} have achieved excellent quality prediction performance on difficult UGC IQA/VQA tasks. Due to their ability to learn complex, hierarchical visual features, they are able to capture complex interactions  between distortions and visual content. More recently, Vision Transformers (ViTs) and Vision-Language Models (VLMs) such as CLIP-IQA~\cite{wang2023exploring}, BLIP~\cite{li2022blip} and viCLIP~\cite{wanginternvid} have been trained to conduct IQA and VQA tasks. It was recently found that simply computing the cosine distance between the intermediate features of ViT-based foundation models like CLIP~\cite{CLIP} and DINOv1~\cite{caron2021emerging} outperforms trained IQA models in accuracy and robustness to geometric distortions, even without fine-tuning on quality assessment datasets~\cite{ghildyal2024foundation}. ViTs apply transformer architectures, which can more effectively capture long-range dependencies and global context on images and videos. VLMs, which integrate visual and textual information, offer a multi-modal perspective, which may enhance the ability of quality prediction models in more complex and contextually rich scenarios. These advancements suggest the potential of deep learning and transformer-based models to offer more robust and adaptive quality assessment capabilities.

\begin{table*}[t!]
\centering
\caption{Summary of AI generated Image and Video Datasets.}
\label{table:datsets}
\begin{tabular}{|l|c|c|c|c|c|c|}
\hline
\textbf{Database} & \textbf{Year} & \textbf{Models} & \textbf{Category} & \textbf{Dataset Size} & \textbf{Data Type} & \textbf{Ratings}                    \\ \hline
HPD~\cite{HPD}               & 2023          & 1               & T2I               & 98,807                & Image              & \multicolumn{1}{l|}{Coarse-grained} \\ \hline
ImageReward~\cite{imagereward}       & 2023          & 3               & T2I               & 136892                & Image              & \multicolumn{1}{l|}{Coarse-grained} \\ \hline
Pick-A-Pic~\cite{pickapic}        & 2023          & 6               & T2I               & 500000                & Image              & \multicolumn{1}{l|}{Coarse-grained} \\ \hline
AGIQA-1K~\cite{AGIQA-1K}          & 2024          & 2               & T2I               & 1080                  & Image              & Fine-grained                        \\ \hline
AGIQA-3K~\cite{AGIQA-3K}          & 2024          & 6               & T2I               & 2982                  & Image              & Fine-grained                        \\ \hline
AIGCIQA~\cite{AIGCIQA2023}      & 2024          & 6               & T2I               & 2400                  & Image              & Fine-grained                        \\ \hline
AIGIQA-20K~\cite{AIGIQA-20K}        & 2024          & 15              & T2I               & 20000                 & Image              & Fine-grained                        \\ \hline
PKU-AIGIQA-4K~\cite{PKU-AIGIQA-4K}     & 2024          & 3               & T2I and I2I       & 4000                  & Image              & Fine-grained                        \\ \hline
AGIN~\cite{AGIN}              & 2024          & 18              & Mixed           & 6049                  & Image              & Fine-grained                        \\ \hline

DCU~\cite{DCU-Chivileva}               & 2023          & 5               & T2V               & 1005                  & Video              & Fine-grained                        \\ \hline
VBench~\cite{vbench}            & 2023          & 7               & T2V               & 3500                  & Video              & Fine-grained                        \\ \hline
FETV~\cite{FETV}              & 2023          & 4               & T2V               & 6984                  & Video              & Fine-grained                        \\ \hline
EvalCrafter~\cite{evalcrafter}       & 2024          & 4               & T2V               & 2476                  & Video              & Fine-grained                        \\ \hline
T2VQA-DB~\cite{t2vqa-db}   & 2024          & 9        & T2V               & 10000                 & Video              & Fine-grained                        \\ \hline
AIGCBench~\cite{aigcbench}         & 2024          & 5               & I2V               & 3950*                 & Video              & Fine-grained                        \\ \hline
\end{tabular}
\end{table*}

\subsubsection{IQA and VQA Models for AIGC}\label{sec:iqa_vqa_aigc}
Over the past decade, significant efforts have been made to develop models that can measure the quality of GenAI images. The Inception Score (IS) uses the Inception V3 image classification model to evaluate how well synthesized images are classified as one of 1,000 known objects~\cite{salimans2016improved}. This score combines the confidence of conditional class predictions (quality) with the marginal probability of the predicted classes (diversity). The Frechet Inception Distance (FID) also uses the Inception V3 network but compares the distributions of generated images against a distribution of real images ("ground truth")~\cite{heusel2017gans}. The Frechet Video Distance (FVD) extends this concept to video data, incorporating the temporal dimension using a 3D convolutional network (e.g., I3D - Inflated 3D ConvNet)~\cite{unterthiner2018towards}. Various improved models include HYPE~\cite{hype}, Clean-FID~\cite{parmar2021cleanfid}, Kernel Inception Distance (KID)~\cite{binkowski2018demystifying}, and CLIP Maximum Mean Discrepancy (CMMD)~\cite{jayasumana2024rethinking} claim to address FID's shortcomings. FVD has been less studied; however recent research indicates it is less sensitive to temporal information than expected, suggesting a need for further investigation~\cite{ge2024content}.

While distribution-based models can evaluate AI-generated content (AIGC) against the distributions of real images, they often do not align well with human preference scores and may not effectively judge the criteria that is specified in text prompts. In this context, real images refer to those created by artists or captured by photographers. Thus, while these metrics are valuable for understanding and validating distribution similarities, they mainly assess whether images contain natural content. Therefore, we also need metrics able to assess the degree of alignment between a user's prompt and the consequent generated image. To explain this further, revisit Fig.~\ref{fig:common_sense}, which shows four AI images generated using the same prompt. During or just after the image has been generated, one needs to verify whether ``the Mandalorian has a weathered backpack slung over his shoulders.'' However, as evident from the four exemplar outcomes shown in the Figure, there remain significant issues with the alignment of the generated image with respect to the provided input prompt.

One metric which can help improve assessment is the CLIP score, which aligns visual and textual representations within a shared embedding space, thereby imparting the ability to effectively understand and connect images with corresponding text descriptions. Variants of the CLIP score, such as BLIP~\cite{li2022blip} and viCLIP~\cite{wanginternvid}, and the metrics R-Precision~\cite{xu2018attngan} and CLIP-R-Precision~\cite{park2021benchmark}, which use the generated image to retrieve the top text matches from 100 text candidates, have been shown to perform well on such tasks.

However, recent research has revealed discrepancies between CLIP scores, FID scores, and human preferences~\cite{HPD}. Consequently, a new model called HPS was proposed, trained on the human preference labels~\cite{HPD} of an AI-generated content dataset. Similarly, the PickScore~\cite{pickapic} model uses CLIP, trained on a large dataset containing human preferences. The ImageReward~\cite{imagereward} model employs reinforcement learning to predict human preferences of AI-generated content, assessing it according to three criteria: \textit{alignment with text}, \textit{realism}, and \textit{safety}. These measurements can also be used to optimize diffusion models, resulting in outputs having improved quality. The evaluation of text-to-video generation is a topic that has been less explored, with current assessments relying on text-to-image metrics. Benchmarks like VBench~\cite{vbench} and FETV~\cite{FETV} have made advances in evaluating videos across a broader range of criteria than those used in text-to-image metrics. However, these studies do not propose a single comprehensive metric for all criteria, but instead utilize a combination of existing metrics. 

Very recently, a unified metric has been introduced, trained on a Large-scale Generated Video Quality (LGVQ) dataset, that can assess videos across multiple criteria~\cite{zhang2024benchmarking}. However, metrics trained on human preferences often prioritize image aesthetics over alignment with user-provided prompts~\cite{HPD}. Moreover, while metrics exist to assess aesthetics and alignment with prompts, semantic considerations and common sense are sometimes overlooked. Examples of efficiencies in common sense are evident in Figure~\ref{fig:common_sense}, where aspects like physical plausibility, general knowledge, and consistency across generated images are compromised. Therefore, one need metrics that can evaluate such criteria more effectively.

\subsection{Subjective Image and Video Quality Assessment}
Subjective quality assessment refers to experiments to collect the “opinions” (ratings) of multiple subjects about the performance of a system for different numbers of well-defined test conditions~\cite{p800_2}. Since the end goal of QoE assessment is to estimate the visual experiences of humans, subjective assessment methods are considered to be the ground truth of the QoE of a service, often reported as the average of the opinion scores termed ``mean opinion scores" (MOS). These assessments are usually conducted in controlled (lab) environments with guidelines outlined in ITU-T Rec. 910~\cite{p910} for multimedia applications, ITU-T Rec. 919~\cite{p919} for 360º video on head-mounted displays, and ITU-R BT.500-15~\cite{bt500-15} for television images. For example, ITU-T Rec. P.910, first published in 1996, and updated recently in Oct, 2023, describes non-interactive subjective assessment methods for evaluating the one-way overall video quality for multimedia applications \cite{p910}. The recommendation discusses selection of test materials (recording environment and system, scene characteristics in terms of spatial and temporal complexity), test methods (ACR, DCR, etc.) and experimental design (complete randomized design; Latin, Graeco-Latin and Youden square designs; and replicated block designs); evaluation procedures (viewing conditions, processing and playback system, selecting viewers, etc.) and statistical analysis and reporting of results.

These standards have been developed over the years to ensure the results are consistent, reliable and reproducible under the test conditions. With standardized assessments, one can ensure that every test participant experiences similar conditions, questions, and evaluation criteria, which minimizes variability in responses due to external factors. Such standardized subjective test methodologies enable researchers to compare results across different different labs (demographics), facilitating broader insights and more robust conclusions. Another significant advantage of standardized subjective assessment methodologies is reproducibility. 

Over past two decades, many subjective image and video quality assessment datasets~\cite{LIVEVideoQualityAssessmentDatabase,BVI_CC,GamingVideoSET,LiveMobileStallVideoDatabaseII,LIVENetflixVideoQualityofExperienceDatabase,LFOVIAVideoQoEDatabase,barmanMultiScreen,avt-uhd1,hosu2024uhdiqa,koniq10k,ying2020patches,ying2021patch}, have been created, most largely following recommended practice regarding subject training, stimulus timing and presentation. Many of these are open-source datasets, with reproducible results that have been an important factor in ensuring the growth of the field of image and video quality assessment. Of course, while standardized subjective assessment methodologies have contributed to the development of best practices and benchmarks in the field of visual quality assessment, the field of visual quality is very fast-changing, and new technologies arise for which new human studies are required. As videos become denser in pixel counts, frame rate, pixel depths, and more immersive, researchers often need to design their own protocols where ITU recommendations fail to apply or may need modification. Nevertheless, by adhering to these standards as best possible, researchers can help contribute to a growing body of knowledge that is coherent and comparable, leading to better outcomes and innovations.

\begin{figure}[t]
    \scriptsize
    \centering
      \includegraphics[width=0.6\linewidth]{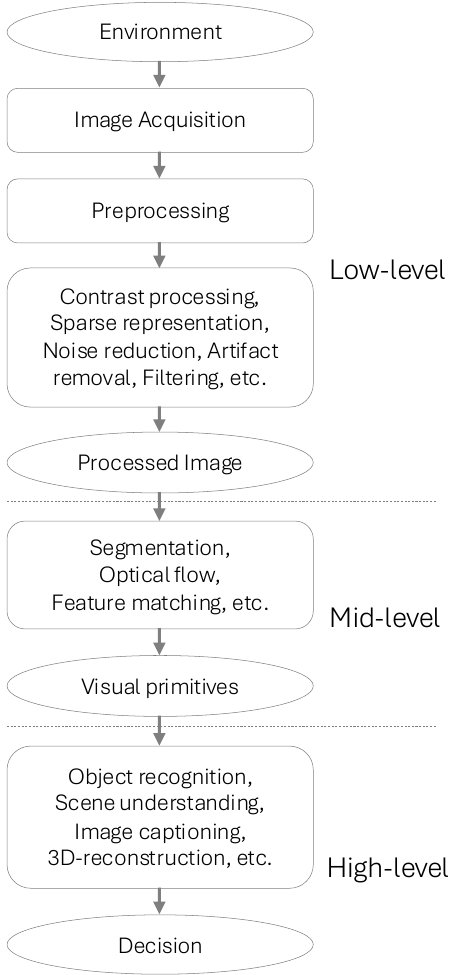}
      \vspace{0.1in}
    \caption{Computer vision spans multiple levels of analysis, from low-level tasks such as noise reduction to mid-level tasks like color segmentation, and finally to high-level tasks that involve interpreting complex interactions, behaviors, and contextual understanding in visual data.}
    \label{fig:levels_cv}
    \vspace{-0.1in}
\end{figure}

\subsubsection{Subjective AIGC Datasets} \label{sec:subjective_aigc_datasets}

Over the past few months, there have been significant efforts towards the creation of open-source AI-generated image and video datasets along with subjective ratings and comparisons of quality models. Table~\ref{table:datsets} presents an overview of these valuable data resources. However, existing AIGC subjective datasets suffer from various shortcomings that could undermine the reliability, validity, and generalizability of objective IQA and VQA models that are trained and evaluated on them, as we discuss next.

One primary issue is the lack of adequate numbers of human raters that subjectively annotate the datasets. As shown in Table~\ref{table:datsets}, early AIGC datasets included single-subject human quality ratings of each video. A database that includes inadequate numbers of subjectively labeled data samples can lead to inconsistencies and biases when building models.

\section{Challenges and Opportunities} \label{sec:challenges}

In this section, we present some of the most interesting challengesin the pursuit of IQA and VQA modeling for AIGC and AIEC.

\subsection{Objective and Subjective Quality Assessment of AI Generated and Enhanced Content}

\subsubsection{Subjective Assessment}

In Section~\ref{sec:subjective_aigc_datasets} we discussed some shortcomings of existing AIGC IQA and VQA subjective datasets. One shortcoming is the current lack of standard methodologies for the design and execution of subjective studies of GenAI content quality. While there are standards for image and video quality assessment, they have been designed to guide studies of the quality of natural and screen content, and using the same protocols without modification may prevent researchers from being able to capture the nuances of AI generated and enhanced content. 

The existing ITU-T and ITU-R recommendations include multiple subjective assessment methodologies: Absolute Category Rating (ACR), Absolute Category Rating with Hidden Reference (ACR-HR), Double Stimulus Continuous Quality Scale (DSCQS), Degradation Category Rating (DCR), and Paired Comparison (PC). Each test methodology (ACR, ACR-HR, DSCQS, DCR, and 2AFC) is generally suited for specific application contexts. While ACR and ACR-HR are suitable for subjects viewing large numbers of contents having wide ranges of distortions, DCR/DSIS and DSCQS are appropriate on contents afflicted with only subtle distortions that may only be observed by comparison against a reference. 2AFC is also used when subjects compare quality differences, including comparing distorted contents to other distorted (same) contents. Alternative methods like SSCQE and Paired Comparison provide high precision at the cost of increased complexity. 

However, as discussed previously, the ``quality” of AI-generated visual content often has different meaning than ``natural'' visual content, and hence GenAI subjective dataset and quality prediction models should reflect this. Future subjective tests of GenAI image and video quality will need to consider multiple aspects of viewing experience beyond technical quality. New studies will need to find ways to probe viewers' opinions of naturalness, aesthetics, and even ``creepiness,'' as GenAI can easily lead to ``uncanny valley'' concerns~\cite{mori2012uncanny}. Moreover, there will be need to be ways to account for artistic intent, and human raters may need to consider the prompts used to generate content. After all, 6-legged horses and pink elephants bowling might be fun to watch! These kinds of considerations point up the extreme difficulty in designing study protocols for measuring subjective impressions of GenAI content, and the even greater difficulty of creating standardized recommendations. GenAI is evolving rapidly in many surprising directions, which may quickly render standard recommendations for study designs obsolete.

\begin{figure*}[t]
    \scriptsize
    \centering
      \includegraphics[width=\linewidth]{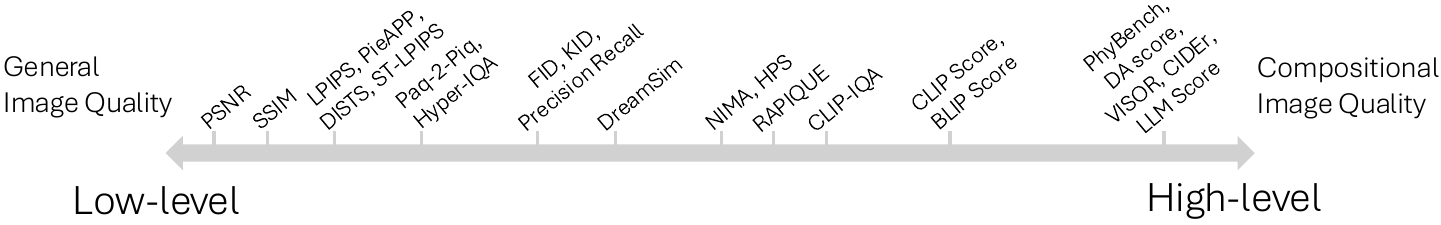}
    \caption{IQA and VQA models represented across a spectrum from low to high levels of abstraction of perceptual quality and/or similarity.}
    \label{fig:metric_levels}
\end{figure*}

\subsubsection{Objective Assessment}

In Section~\ref{sec:objective_iqa_vqa}, we outlined recent efforts on designing objective image and video quality prediction models for measuring the visual quality of AIGC. Existing models that only assess attributes like aesthetics, statistical naturalness, and fidelity, are inadequate to analyze GenAI content, since they fail to address other relevant attributes, and perform poorly on GenAI videos and pictures~\cite{aigcbench}. However, advancements in Vision-Language Models (VLMs) offer the promise of viable, general solutions~\cite{radford2021learning}. It is clear that the evaluation criteria for AIGC need to extend beyond just traditional technical quality; other relevant dimensions, such as geometric, structural, and even biological consistency and visual realism must be considered. These factors include various elements that may be inferred by modern computer vision models, such as lighting, color, texture, shape, and motion, all of which impact the interpretation of and quality of images and videos, as demonstrated by recent works encompassing multiple quality attributes~\cite{lee2024holistic, AGIQA-1K, vbench, bansal2024videophy, meng2024phybench}. 

In a recent work on evaluating AIGC, a taxonomy of models was proposed, by categorizing them into general image quality predictors and compositional image quality analyzers~\cite{hartwig2024evaluating}. General image quality evaluation focuses on assessing realism, distortions, artifacts, and aesthetics, while compositional image quality focuses on attribute binding, spatial relationships, and object accuracy. We propose a more continuous scale of visual quality models based on a hierarchy of perceptual similarity measurements. As shown in Fig.~\ref{fig:metric_levels}, on this scale models like LPIPS, PieAPP, DISTS, ST-LPIPS, and PSNR focus on low-level perceptual similarity or visual fidelity features. Distribution-based models, such as FID, capture broader statistical similarities than PSNR or LPIPS~\cite{zhang2018unreasonable}. NIMA~\cite{talebi2018nima} targets both aesthetics and technical quality. RAPIQUE~\cite{Rapique} separately measures technical and semantic attributes of video quality thereby mapping both low- and high-level features to predict human impressions of quality. CLIP-IQA~\cite{wang2023exploring} judges technical picture quality while also conducting abstract quality assessment based on attributes such as complex/simple, new/old, scary/peaceful, etc. LLMScore~\cite{lu2024llmscore}, DA Score~\cite{singh2023divide}, PhyBench~\cite{meng2024phybench}, VISOR~\cite{gokhale2022benchmarking}, and CIDEr~\cite{vedantam2015cider} focus on mid-to-high-level common sense and geometric features. Categorizing GenAI IQA/VQA models on a low-to-high perceptual similarity scale better identifies how they are associated with GenAI image and video quality evaluation and how complex the model architectures may be.

\subsubsection{Enhancing Commonsense Understanding in IQA/VQA}
In Section~\ref{sec:iqa_vqa_aigc}, we explored various models that assess human preferences across multiple criteria. However, when attempting to build models that evaluate common sense understanding of images such as the aspects exemplified in Fig.~\ref{fig:common_sense} human opinion scores, even taken in larger numbers, may not alone suffice. Conversely, benchmarks like Holistic Evaluation of text-to-image models (HEIM)~\cite{lee2024holistic}, which are an initial but limited effort to address this complex issue, necessarily fall short since they rely exclusively on CLIP to assess common sensical aspects of analyzed images. Thus, human impressions of quality are excluded. No doubt, more advanced models will be devised in the future that span different levels of abstraction of perceptual quality and similarity. To enable this, new insights will be required to design and create subjective datasets that encompass hierarchies of quality abstraction.

\subsection{Descriptive Quality Assessment}

More recently, there has been an increasing use of Large Multimodal Models (LMMs) to evaluate image quality~\cite{wu2023q, huang2024visualcritic, you2023depicting, ku2023viescore, huang2023t2i, wang2024large}. These systems have the potential to be able to provide detailed descriptions of the various types of distortions that are present in images. This descriptive information could serve as the foundation more sophisticated scoring mechanisms that capture nuanced aspects of image quality, beyond the simple quality scores predicted by current models. The challenges will include harnessing these detailed descriptions to devise scoring systems that more accurately reflect human subjective impressions of image and video quality.

However, LLMs still require significant improvement as zero-shot predictors of visual quality in the full-reference IQA and VQA settings~\cite{zhu20242afc}, indicating that much work remains to be done. Recent advancements in this direction include newer frameworks using Chain of Thought (CoT) prompting~\cite{wei2022chain} which relies on a comprehensive set of intermediate logical explanations linked together in sequence, making it useful for complex reasoning. Several recent methods have incorporated this into the evaluation methodology of the respective IQA models~\cite{chen2023x, huang2023t2i} which has been shown to be effective for different types of distortions. Given the complexity of perceptual reasoning involved in quality assessment, such frameworks could be highly beneficial. Moreover, such models when combined with other evaluation aspects, such as geometric and semantic information, can lead to better explainable IQA models~\cite{wu2024comprehensive}.

\subsection{Local Patch Scores}

We have observed that learned image and video quality assessment methods can be better predictors of human judgments of perceptual quality compared to traditional methods, when there is adequate perceptual data. However, understanding the specific spatial and/or elements of the quality-analyzed images or videos that most heavily influence the final quality score produced by IQA/VQA models remains challenging. Recent work towards addressing this gap involves local patch quality analysis and prediction. For example, PaQ2PiQ~\cite{ying2020patches} provides local patch quality measurements, allowing determination of where the worst perceptual errors occur. The same authors extend patch-based quality analysis on the spatial, temporal, and space-time patches of videos~\cite{ying2021patch}. These approaches can be used to effectively in explain how different parts of images and videos contribute to final quality predictions.

\subsection{AI Safety: Legal and Ethical Aspects}

AI safety is a major concern for computer vision and visual communication systems, especially as they are increasingly employed in safety-critical applications and multimedia content generation. Since IQA and VQA can be integral parts of these systems they are also vulnerable to adversarial attacks, but they can also help identify these issues. First, we will discuss how to ensure adversarial robustness of IQA/VQA models.

Recent studies have revealed that IQA models are susceptible to attacks via the addition of adversarial imperceptible perturbations~\cite{zhang2022perceptual, NEURIPS2023_a1c71663, shumitskaya2024towards, shumitskaya2024ioi, antsiferova2024comparing, leonenkova2024tipatch, ghildyal2023,ghazanfari2023r, kettunen2019lpips, gushchin2024guardians}. These perturbations can drastically alter the quality predictions produced by these algorithms. Unlike natural perturbations, these are deliberately designed by an attacker to exploit weaknesses of the models. For full-reference models, these perturbations can be transferred from one model to another~\cite{ghildyal2023}, meaning that adversarial perturbations designed for a publicly available IQA model can also be used to attack a different model whose internal details are unknown to the attacker. Hence, developing adversarially robust IQA models is important. Such models can provide additional advantages, as demonstrated by the adversarially robust CLIP model for IQA~\cite{croce2024adversarially}, which not only surpasses the performance of existing models but can also detect Not-Safe-for-Work (NSFW) content. Recent work by Leonenkova \textit{et al.}~\cite{leonenkova2024tipatch} demonstrated that NR-IQA models are vulnerable to patch attacks. Adversarial patches can mislead visual quality measurement systems into perceiving pictures or videos as being of high quality when they are not, potentially resulting in increased streaming costs if the video transcoding and subsequent operations depend on the measured video quality. Currently, the potential vulnerabilities caused by adversarial attacks on IQA models have not yet been fully explored. However, recent developments in research on the adversarial vulnerabilities of these models highlights the need for robust defense methods to prevent such attacks. As adversarial attacks continue to advance, it will be crucial to develop and implement stronger defense mechanisms. Since emerging IQA/VQA models are being developed for AI-generated content, these vulnerabilities will likely afflict GenAI quality predictions models as well.

\section{Emerging Trends in IQA and VQA} \label{sec:trends}

\subsection{Fusion-based Approaches}
We have observed a new trend in models that ensemble multiple pre-trained IQA/VQA models to enhance generalization and accuracy for quality assessment tasks. Recent models, such as COVER~\cite{cover}, demonstrate significant improvements by employing a three-branch architecture, where each branch is responsible for capturing different levels of visual quality analysis tasks: low, medium, and high. For example, COVER utilizes CLIP~\cite{CLIP} for high-level visual analysis, a trained CNN for medium-level quality tasks, and a technical IQA-trained model (addressing noisy content) for low-level quality analysis. This approach allows the model to generalize better when facing new application domains and distortion types. Although these models suffer have a large number of parameters due to merging multiple large vision architectures, they exhibit very high performance.

\subsection{Dataset Creation and Psychovisual Research}
Apart from using AIGC for multimedia purposes, it can also be used to collect data for psychovisual experiments. DreamSim learns from both mid-level and low-level perceptual features, making it valuable for both image retrieval and evaluating distortions generated by image reconstruction methods~\cite{fu2024dreamsim}. Curating a dataset to train such a model is challenging; therefore, the authors of DreamSim utilized stable diffusion~\cite{rombach2022high} to generate relevant images for comparison. In the future, we expect more datasets to be created using newer text-to-image and text-to-video methods. 

\subsection{Beyond Geometric Consistency and Physical Fidelity}

Because capturing the nuances of physical laws in AI models and datasets is challenging, AIGC models often produce unrealistic scenes. A recent study introduced a method for evaluating the ability of video generation methods to maintain geometric consistency across contiguous video frames~\cite{li2024sora}, an aspect traditional models like FID~\cite{heusel2017gans} and FVD~\cite{unterthiner2018towards} struggle to achieve. One possible approach is to leverage 3D reconstruction and geometric feature matching, to determine whether geometric features remain consistent across frames and retain the same geometric structures over prolonged durations~\cite{li2024sora}. However, beyond geometric consistency, AIGC must also embody physical realism and fidelity. Assessing the accuracy by which a generative AI model replicates real-world physical principles and behaviors requires models that can verify the alignments of object motions in videos against real-world counterparts. The only model are aware of for this purpose is VideoPhy~\cite{bansal2024videophy}, which uses an LLM to predict semantic adherence and physical commonsense in videos. VideoPhy is fine-tuned on human annotations to validate alignment with captions. 

\subsection{Physical Commonsense and Material Perception}
AIGC often does not adhere to the physical laws of the real world. PhyBench is a recent study that explores the evaluation of physical commonsense in models across four aspects of physical knowledge: material properties, mechanics, thermodynamics, and optics~\cite{meng2024phybench}. This approach could be enhanced by integrating additional aspects of material perception. These might be derived from existing studies of material perception~\cite{schmidtcore, filip2024perceptual}. In a study focused on utilizing vision-language models for material perception, the authors argued that constructing a comprehensive evaluator would require incorporating not only other modalities, such as language, but also leveraging visual features to effectively differentiate materials~\cite{liao2024probing}. 

\subsection{Leveraging AI Image Detection Strategies to Uncover Geometric Inconsistencies in AIGC}

AI image detection and GenAI for IQA both involve analyzing GenAI images and share similarities in their methodologies. They often use comparable feature extractors, datasets, and performance metrics, such as correlation with human perception. Therefore, methods that analyze geometric inconsistencies to detect AI-generated images might also be used to develop IQA models. For example, a recent study focused on distinguishing AI-generated face images from real ones, by constructing 3D models from the images~\cite{bohavcek2023geometric}. These geometric models capture structural properties of the faces, then features derived from them are used to build a classifier. Another recent study utilized discrepancies in the geometry of generated images to detect AI-generated content~\cite{sarkar2023shadows}. This study specifically analyzed inconsistencies in object shadows through instance detection, and assessed perspective geometry using a Line Segment Detector, perspective fields, and point cloud processing with deep learning. Building on these approaches, future research could aim towards developing models of physical fidelity that measures the realism of both object appearances and of the scenes they are placed in.

\subsection{Sustainability}
Training and deploying large AI models demand significant computational resources. Beyond the complexity of the models, the efficiency of the hardware used, such as GPUs and TPUs, plays a crucial role in determining the overall carbon footprint. Recently, considerable attention has been directed towards examining the interplay between power consumption and video quality in streaming contexts~\cite{katsenou2024rate, 10219577, 10077211, beuscart2023listening, 10566286, 10447099}. Given the increasing emphasis on the relationship between power consumption and video quality in streaming contexts, it is similarly critical to evaluate how power-saving methods impact visual quality in emerging technologies. In a recent study, `Perceptual Evaluation of Algorithms for Power Optimization in XR Displays (PEA-PODs)~\cite{chen2024pea}, the authors emphasized the importance of assessing how power-saving methods impact visual quality. This study used perceptual models and user studies to identify the best image mapping techniques to extend battery life in virtual extended reality (XR) displays. Similar practical studies on power consumption may prove crucial in the perceptual evaluation of AIGC. Additionally, developing ways of assessing the power consumption of AIGC is equally important. Several authors have noted that the requirements of high computational power to sustain AI models have been increasing~\cite{lin2024exploding, li2023making, zuccon2023beyond, li2023toward}. Given this growing demand for computational power, there is a need to accelerate improvements in AI model efficiency. Creating generative AI models requires substantial energy, resulting in significant carbon emissions and water consumption for cooling. Therefore, it is crucial to evaluate the sustainability of generative AI to justify its environmental impact and to efficiently employ generative AI tools.

\subsection{Evaluation of Novel View Synthesis}
Current approaches for comparing novel view synthesis (NVS) methods typically involve computing image quality via using algorithms like PSNR, SSIM, and LPIPS on a subset of hold-out views for a few scenes. The main objective of NVS methods is to accurately render a scene or object of interest from a particular viewing angle desired by the user. 

\textit{Temporal artifacts.} As the user continuously interacts with a scene, maintaining a continuous representation of the scene necessitates understanding the physics governing light changes and the movements of objects from various angles. However, since NVS models neither explicitly validate nor incorporate indicators of the physical plausibility of the reconstructed environment, continuous viewing of the scene may include flickering and ghosting artifacts, similar to those sometimes observed in video reconstruction methods. Hence, it may be that the evaluation of quality should be conducted using videos instead of static views~\cite{Liang2024,hou2022vfips}. Currently, popular approaches for evaluation rely on IQA models that fail to account for temporal artifacts.   

\textit{Unseen distortions.} In a recent study, it has been observed that commonly used learned algorithms like LPIPS struggle to accurately assess the quality of NeRFs. A primary reason for this lack of generalization are the differences in the distributions of NeRF generated content and what content the IQA models were trained on~\cite{Liang2024}. However, another study reaches a contrasting conclusion, suggesting that learned IQA models perform well at evaluating NeRFs despite not being specifically trained for that task~\cite{martin2024nerf}. As advancements continue in the area of NVS, for example Gaussian Splatting-based approaches~\cite{kerbl20233d}, it will be important to investigate whether established IQA models can effectively assess the performances of newer NVS methods.

\textit{No-reference NeRF IQA.} The absence of reference images or videos complicates the evaluation of NVS methods. Recent work on no-reference NeRF quality assessment demonstrates that a well-crafted model can outperform the accuracy of even FR-IQA algorithms~\cite{qu2024nerf}. A reduced reference model~\cite{rehman2012reduced} or a near-aligned reference model~\cite{liang2016image} can be effective, by using relevant features computed from images taken from a neighboring viewpoint of the one used to capture the original content for comparison.

\textit{Text-to-3D.} Several methods utilize diffusion models for text-to-3D synthesis. These approaches often optimize a NeRF with a diffusion prior to generate images based on the provided text~\cite{pooledreamfusion}. To evaluate these models against other methods, FR-IQA models are often used, although some authors prefer user studies~\cite{lin2023magic3d}.

\section{Conclusion} \label{sec:conc}

This paper has aimed to conduct a comprehensive overview of the field of visual quality assessment of AI-generated and AI-enhanced content. Significant progress has been made in the fields of AI-generated and AI-enhanced content (AIGC and AIEC) over the past decade, and these technologies are gradually becoming integrated into mainstream applications. This growing enthusiasm brings not only exciting challenges and opportunities but also a responsibility to collaborate on standardizing testing methodologies and benchmarks, as well as developing more accurate and efficient models for GenAI Image Quality Assessment (IQA) and Video Quality Assessment (VQA). One major challenge and opportunity is the development of explainable models that can interpret quality scores and apply them in real-world systems for Quality of Experience (QoE) modeling, monitoring, and optimization. Another critical challenge is the establishment of common benchmarks and standardized subjective testing protocols for GenAI content, encompassing aspects such as test content selection, testing methodologies, display settings, and other relevant parameters. As this field continues to evolve, it is crucial that the research community works together to ensure that these advancements are implemented in ways that are innovative, reliable, and responsible.

% \begingroup

% \setlength\bibitemsep{0pt}
% \printbibliography
% \endgroup
\bibliographystyle{IEEEtran}
\bibliography{00_main}

\begin{IEEEbiography}
[{\includegraphics[width=1in,height=1.25in,clip,keepaspectratio]{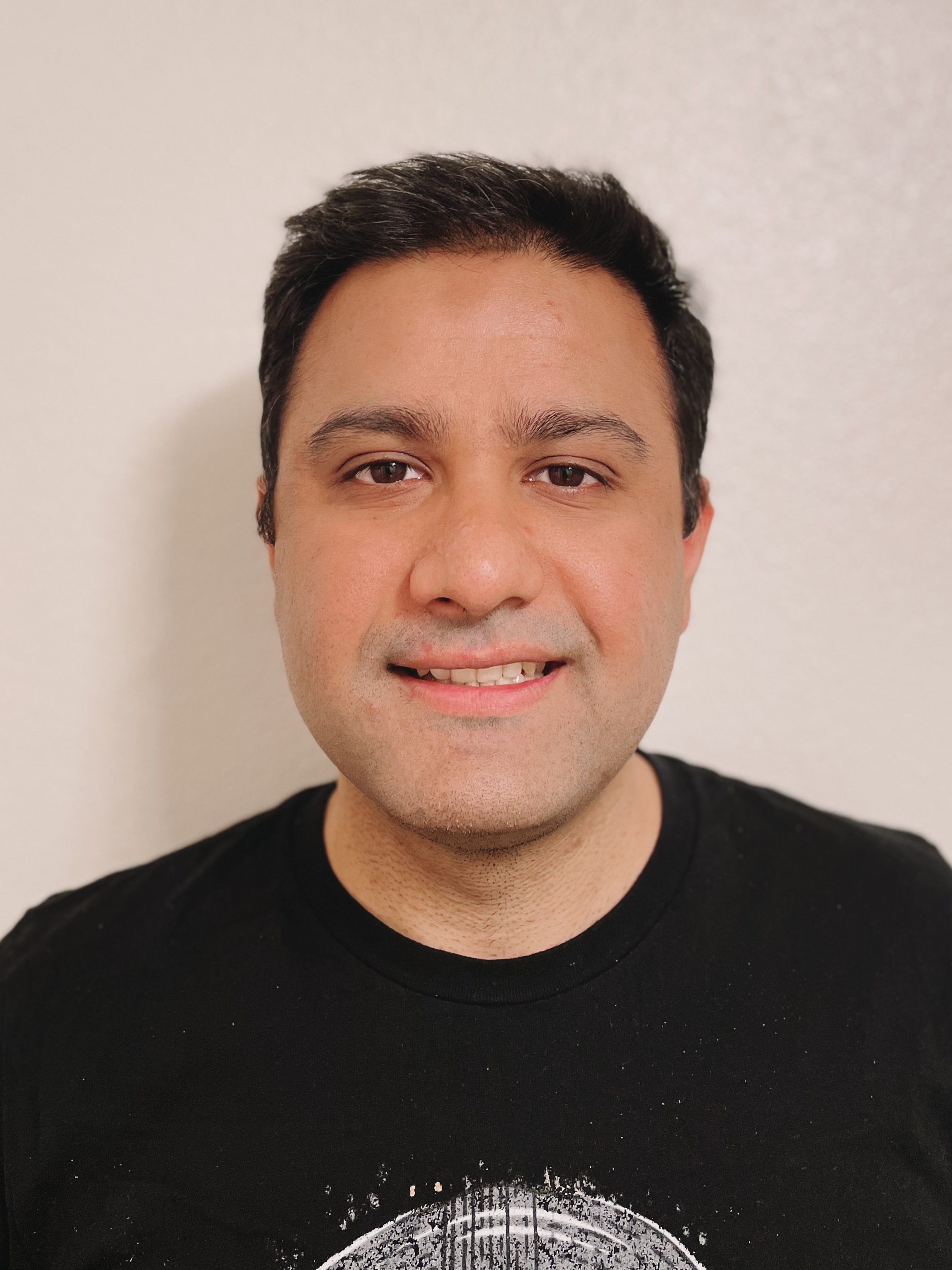}}]{Abhijay Ghildyal} is a final-year Ph.D. student in Computer Science at Portland State University’s Computer Graphics and Vision Lab, advised by Dr. Feng Liu. He specializes in low-level computer vision and deep learning. His work has been presented at prominent venues, covering topics such as perceptual similarity metrics and video frame interpolation. Abhijay has held various research positions in the industry at companies such as Amazon, Mu Sigma Inc., Cambia Health Solutions, and Quantela Inc. He is also the recipient of the Richard Kieburtz Memorial Graduate Fellowship. His research interests broadly lie in image and video quality assessment, perceptual similarity metrics, computational visual perception, novel view synthesis, machine learning safety, adversarial robustness, and precision agriculture. Abhijay has published in leading international conferences and journals and serves as a reviewer for several academic venues.
\end{IEEEbiography}

\begin{IEEEbiography}
[{\includegraphics[width=1in,height=1.25in,clip,keepaspectratio]{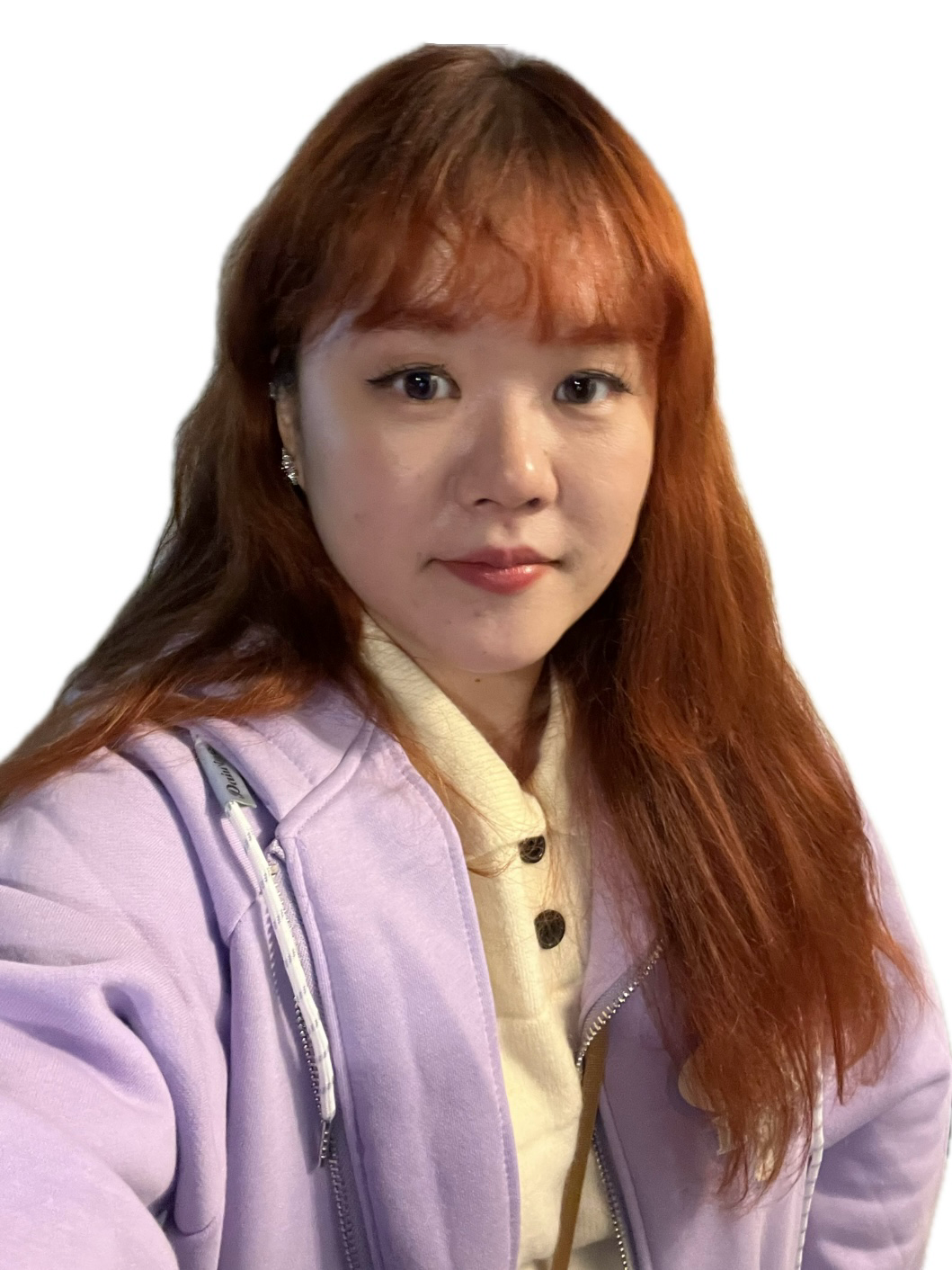}}]{Yuanhan Chen} is a Data Scientist in the Future Technology Group at Sony Interactive Entertainment (PlayStation). She is a researcher specializing in the development of Quality of Experience (QoE) models for cloud gaming and game streaming platforms. With a focus on enhancing user interaction and satisfaction, her work integrates cutting-edge technologies to optimize gaming experiences. She collaborates closely with industry production teams to implement machine learning inference and streamline data pipelines, ensuring efficient and scalable solutions. Additionally, her team is at the forefront of creating evaluation metrics specifically designed for assessing generative AI content, driving innovation in AI-enhanced gaming environments. Her research bridges academia and industry, contributing to the next generation of gaming technology. Dr.Chen holds a Ph.D and M.Sc in Chemical Engineering from the University of California, Irvine. 
\end{IEEEbiography}

\begin{IEEEbiography}
[{\includegraphics[width=1in,height=1.25in,clip,keepaspectratio]{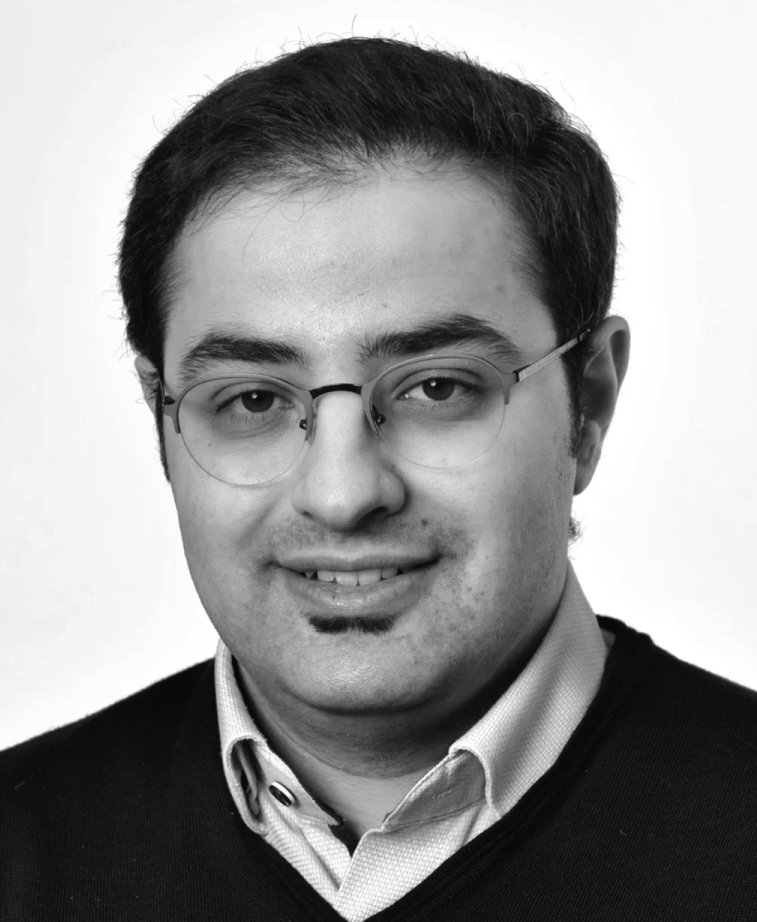}}]{Saman Zadtootaghaj} is a Senior Research Engineer at PlayStation, Sony. His primary focus is on the subjective and objective quality assessment of computer-generated content. Prior to joining Sony, he worked as a researcher at Dolby Laboratories and Telekom Innovation Laboratories, Deutsche Telekom AG. He earned his Ph.D. from the Quality and Usability Lab at TU Berlin, under the supervision of Prof. Dr.-Ing. Sebastian Möller. Saman currently chairs the Computer-Generated Imagery group at the Video Quality Experts Group (VQEG) and is an active member of ITU-T Study Group 12, where he leads the P.BBQCG work item.
\end{IEEEbiography}

\begin{IEEEbiography}
[{\includegraphics[width=1in,height=1.25in,clip,keepaspectratio]{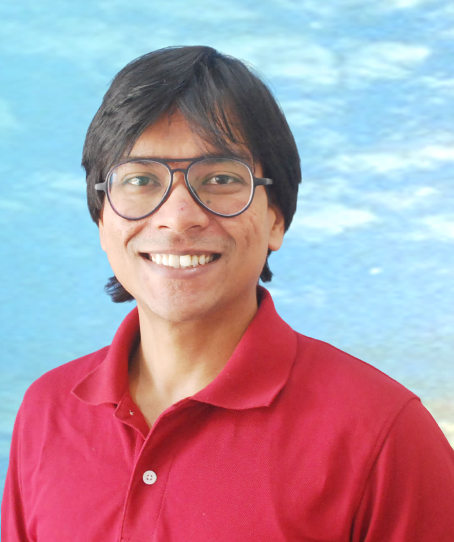}}]{Nabajeet Barman} (SrM'24) is a Senior Research Scientist in the Future Technology Group at Sony Interactive Entertainment (PlayStation) where he is working on the integration of cutting-edge AI algorithms into content-aware encoding strategies and development of quality assessment models for generative AI-enhanced content to enhance user experience on next-generation gaming platforms. Previously, Dr. Barman was a Principal Video Systems Engineer at Brightcove, where he specialized in optimal video encoding strategies, perceptual quality assessment, and end-to-end optimization. He holds a Ph.D. in Computer Science from Kingston University (London, UK) as part of his Marie Curie Fellowship with the MSCA ITN QoE-Net from 2015 to 2018. He also holds an MBA from Kingston University (London, UK), M.Sc. in Information Technology from Universität Stuttgart (Germany), and a B.Tech in Electronics Engineering from the National Institute of Technology, Surat (India). Dr. Barman has also served as a Lecturer in Applied Computer Science (Data Science) at Kingston University, London where he obtained Fellow of the Higher Education Academy (FHEA), UK certification reflecting his commitment to excellence in teaching and learning. From 2012 to 2015, he held various roles across industries, including a tenure at Bell Labs (Stuttgart, Germany). He is a Board Member of the Video Quality Expert Group (VQEG) and an individual committee member of the British Standards Institution (BSI), contributing to the development of standards in multimedia systems. He has published extensively in leading international conferences and journals and is an active reviewer for numerous scholarly publications.
\end{IEEEbiography}

\begin{IEEEbiography}
[{\includegraphics[width=1in,height=1.25in,clip,keepaspectratio]{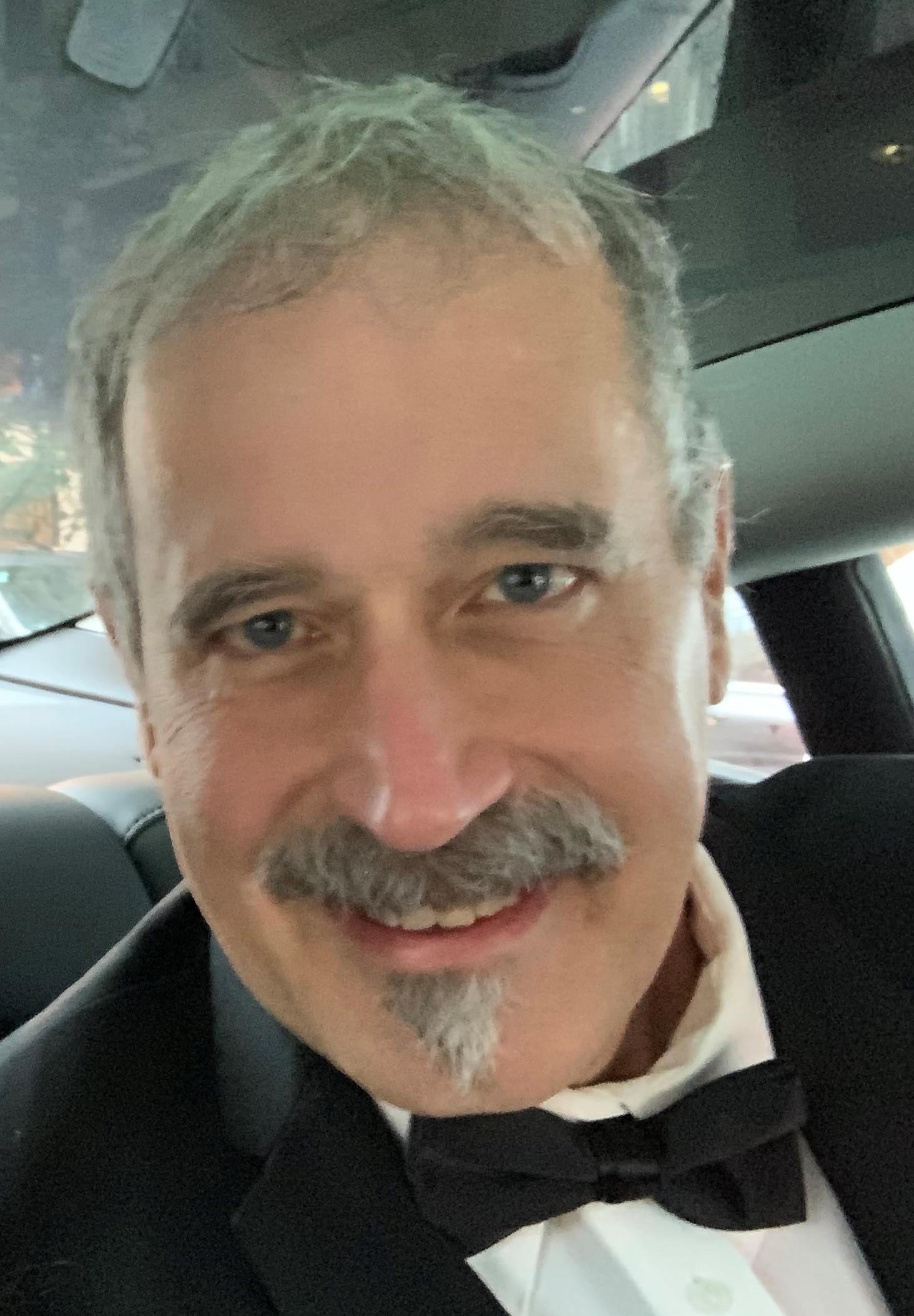}}]{Al Bovik} (LF ‘23) is the Cockrell Family Regents Endowed Chair Professor at The University of Texas at Austin. An elected member of the United States National Academy of Engineering, the Indian National Academy of Engineering, the National Academy of Inventors, and Academy Europaea, his research interests include image processing, computational vision, visual neuroscience, and streaming video and social media. For his work in these areas, he received the IEEE Edison Medal, IEEE Fourier Award, Primetime Emmy Award for Outstanding Achievement in Engineering Development from the Television Academy, Technology and Engineering Emmy Award from the National Academy for Television Arts and Sciences, Progress Medal from The Royal Photographic Society, Edwin H. Land Medal from Optica, and the Norbert Wiener Society Award and Karl Friedrich Gauss Education Award from the IEEE Signal Processing Society. He has also received about 10 ‘best journal paper’ awards, including the IEEE Signal Processing Society Sustained Impact Award. His books include The Essential Guides to Image and Video Processing. He co-founded and was the longest-serving Editor-in-Chief of the IEEE Transactions on Image Processing and created/Chaired the IEEE International Conference on Image Processing which was first held in Austin, Texas, 1994.

\end{IEEEbiography}

\end{document}